\documentclass[runningheads]{llncs}
\usepackage[utf8]{inputenc}
\DeclareUnicodeCharacter{05BF}{} 
\usepackage{eccv}
\usepackage{eccvabbrv}

\usepackage{graphicx}
\usepackage{booktabs}
\usepackage{amsmath,amssymb}
\usepackage{xcolor}
\usepackage{multirow}
\usepackage{pifont}
\newcommand{\cmark}{\ding{51}}
\newcommand{\xmark}{\ding{55}}

\usepackage{hyperref}

\title{When Does a Laugh Begin? Structured Annotator Disagreement in Temporal Laughter Localization}
\titlerunning{When Does a Laugh Begin?}

\author{Eyal~Hanania \and
Daniel~Arkushin \and
Naveh~Ayal \and
Jonathan~Benvenisti \and
Amos~Bercovich \and
Elie~Zemmour\thanks{E.~Zemmour and S.~Froim contributed equally as co-senior authors.} \and
Sahar~Froim\textsuperscript{$\star$}}
\authorrunning{E. Hanania et al.}
\institute{WSC Sports\\
\email{Eyal.Hanania@wsc-sports.com}}

\newcommand{\humanCeiling}{0.681}
\newcommand{\singleGtSwing}{0.246}
\newcommand{\nBenchmark}{672}
\newcommand{\nRaterThree}{449}
\newcommand{\nRaterFour}{211}
\newcommand{\nRaterFive}{12}
\newcommand{\frameAlpha}{0.757}
\newcommand{\frameAlphaSub}{0.757}

\newcommand{\alphaBinLow}{0.756}
\newcommand{\alphaBinHigh}{0.758}
\newcommand{\existUnanimous}{77.2}
\newcommand{\existContested}{22.8}
\newcommand{\onsetSD}{0.146}
\newcommand{\offsetSD}{0.253}
\newcommand{\asymRatio}{1.73}
\newcommand{\asymWilcoxonP}{4.3\times10^{-19}}
\newcommand{\chuckleDisagree}{77}
\newcommand{\fullDisagree}{20}
\newcommand{\chuckleOfContested}{68}
\newcommand{\chuckleOfUnanimous}{14}
\newcommand{\nEvents}{7,950}
\newcommand{\pctAcoustic}{94.3}
\newcommand{\pctBoth}{4.8}
\newcommand{\pctVisual}{0.9}

\newcommand{\pctChuckleIntensity}{24.6}
\newcommand{\pctFullIntensity}{75.4}

\newcommand{\nMatchedEvents}{1,683}
\newcommand{\evPerVideoMean}{2.5}
\newcommand{\evPerMin}{6.24}
\newcommand{\laughShareUnion}{21.1}
\newcommand{\laughShareMaj}{14.7}
\newcommand{\durChuckleMed}{1.16}
\newcommand{\durFullMed}{1.83}

\newcommand{\sgRank}{69.7}
\newcommand{\calRank}{80.0}

\newcommand{\multiRank}{80.4}

\newcommand{\sgRankAnnOrigin}{67.9}
\newcommand{\calRankAnnOrigin}{77.4}
\newcommand{\multiRankAnnOrigin}{77.6}
\newcommand{\bandTiePct}{67}
\newcommand{\bandDecidedAcc}{85}
\newcommand{\fullRank}{79.7}
\newcommand{\rankGapLow}{9}
\newcommand{\rankGapHigh}{11}
\newcommand{\fullRankWLow}{79.5}
\newcommand{\fullRankWHigh}{80.0}
\newcommand{\degBand}{0.248}
\newcommand{\degFull}{0.159}
\newcommand{\worstTierFull}{0.657}
\newcommand{\bandOnsetNinety}{0.5}
\newcommand{\bandOffsetNinety}{0.727}
\newcommand{\bandRatioNinety}{1.45}
\newcommand{\nBandCal}{3,815}
\newcommand{\covOutMean}{0.895}
\newcommand{\covOutLow}{0.86}
\newcommand{\covOutHigh}{0.92}
\newcommand{\adBandOnsetNinety}{0.368}
\newcommand{\adBandOffsetNinety}{0.563}
\newcommand{\bandOnKTwo}{0.535}
\newcommand{\bandOffKTwo}{0.712}
\newcommand{\bandOnKThree}{0.467}
\newcommand{\bandOffKThree}{0.734}
\newcommand{\nBandKTwo}{2,162}
\newcommand{\nBandKThree}{1,653}
\newcommand{\chuckleOddsRatio}{8.59}
\newcommand{\disagreeAUC}{0.831}
\newcommand{\disagreeAUCNoInt}{0.757}
\newcommand{\mixedIntPct}{37.2}
\newcommand{\intAlpha}{0.344}
\newcommand{\chuckleDisagreeUnanInt}{90.2}
\newcommand{\fullDisagreeUnanInt}{23.8}
\newcommand{\asymRatioAud}{1.75}
\newcommand{\asymRatioSpk}{1.54}
\newcommand{\asymRatioSpkLow}{1.06}
\newcommand{\asymRatioSpkHigh}{2.27}
\newcommand{\asymRatioAudLow}{1.56}
\newcommand{\asymRatioAudHigh}{1.97}
\newcommand{\nAsymSpk}{60}
\newcommand{\nAsymAud}{1,328}
\newcommand{\disVisual}{69.0}
\newcommand{\disVisualLow}{62.0}
\newcommand{\disVisualHigh}{75.3}
\newcommand{\nVisualClusters}{184}
\newcommand{\disAcousticOnly}{37.0}
\newcommand{\pctAudienceEv}{94.4}
\newcommand{\nEvAud}{4,298}
\newcommand{\nEvSpk}{256}
\newcommand{\nMarksBench}{4,554}
\newcommand{\chuckleDisagreeAud}{74.4}
\newcommand{\fullDisagreeAud}{19.9}
\newcommand{\chuckleDisagreeSpk}{88}
\newcommand{\fullDisagreeSpk}{41}
\newcommand{\nChSpk}{100}
\newcommand{\nFullSpk}{17}
\newcommand{\chuckleMatchLow}{77}
\newcommand{\chuckleMatchHigh}{82}
\newcommand{\fullMatchLow}{20}
\newcommand{\fullMatchHigh}{25}
\newcommand{\asymRatioLow}{1.3}
\newcommand{\asymRatioHigh}{1.7}
\newcommand{\chuckleDisagreeSub}{75.6}
\newcommand{\fullDisagreeSub}{17.5}
\newcommand{\chuckleGapStratMin}{44}
\newcommand{\chuckleGapStratMax}{55}
\newcommand{\chOnsetSD}{0.175}
\newcommand{\fullOnsetSD}{0.135}
\newcommand{\chOffsetSD}{0.275}
\newcommand{\fullOffsetSD}{0.244}
\newcommand{\nChCom}{395}
\newcommand{\nFullCom}{993}
\newcommand{\chAsymRatio}{1.58}
\newcommand{\fullAsymRatio}{1.81}
\newcommand{\bndOnP}{6.8\times10^{-3}}
\newcommand{\bndOffP}{1.3\times10^{-2}}
\newcommand{\bndQuartRatioLow}{1.6}
\newcommand{\bndQuartRatioHigh}{2.3}
\newcommand{\annKOneRank}{70.6}
\newcommand{\annKTwoRank}{75.6}
\newcommand{\annKThreeRank}{77.4}
\newcommand{\annAllRank}{78.4}
\newcommand{\annFOneLow}{0.6}
\newcommand{\annFOneHigh}{0.78}
\newcommand{\bootAlphaLow}{0.742}
\newcommand{\bootAlphaHigh}{0.769}
\newcommand{\bootGapLow}{52}
\newcommand{\bootGapHigh}{61}
\newcommand{\bootAsymLow}{1.54}
\newcommand{\bootAsymHigh}{1.97}
\newcommand{\decayOverAttack}{1.48}
\newcommand{\decayOverAttackLow}{1.4}
\newcommand{\decayOverAttackHigh}{1.5}
\newcommand{\fracDecayGtAttack}{68}
\newcommand{\nEnvEvents}{1,278}
\newcommand{\rhoDecayOffset}{-0.07}

\newcommand{\proxyAUCsep}{0.727}
\newcommand{\proxyAUCnoint}{0.757}
\newcommand{\proxyAUCperc}{0.831}
\newcommand{\proxyAUCrms}{0.771}

\newcommand{\nAudioVideos}{672}
\newcommand{\blN}{578}
\newcommand{\blNEvents}{1,075}
\newcommand{\blNegVids}{94}
\newcommand{\blNegEvents}{5}
\newcommand{\blSG}{0.503}
\newcommand{\blSwing}{0.308}
\newcommand{\blMulti}{0.501}
\newcommand{\blCal}{0.335}
\newcommand{\blBand}{0.604}
\newcommand{\blCovOnG}{0.829}
\newcommand{\blCovOffG}{0.772}
\newcommand{\blCovOnA}{0.865}
\newcommand{\blCovOffA}{0.825}
\newcommand{\blNMatched}{916}
\newcommand{\blMissContested}{59}
\newcommand{\blMissUnanimous}{31}
\newcommand{\blMissChuckle}{58}
\newcommand{\blMissFull}{33}
\newcommand{\blMissExGap}{28}
\newcommand{\blMissExGapLow}{22}
\newcommand{\blMissExGapHigh}{33}
\newcommand{\blMissIntGap}{25}
\newcommand{\blMissIntGapLow}{19}
\newcommand{\blMissIntGapHigh}{30}
\newcommand{\blRankSG}{57.9}
\newcommand{\blRankMulti}{62.2}
\newcommand{\blRankCal}{63.4}
\newcommand{\blRankFull}{62.6}

\begin{document}
\maketitle

\begin{abstract}
Annotators routinely disagree on laughter boundaries and subtle chuckles, yet temporal laughter localization typically evaluates against a single reference annotation. We show that this disagreement is structured rather than random noise. Re-annotating the SMILE-Temporal benchmark (672 videos, 1,683 events) with 3-5 annotators per video (alpha = 0.757), we find systematic patterns: disagreement is 1.73x larger at offsets than onsets, far more common for chuckles than full laughs (77ֿ\% vs. 20\%), and predictable from event attributes (AUC 0.831). Evaluating against a single annotator breaks down under this structure: system scores shift by 0.246 F1 depending on the chosen ground truth, correctly ranking systems only 69.7\% of the time (vs. ~80\% against all annotators). We propose a disagreement-calibrated evaluation that scores predictions against the full annotator distribution using conformally calibrated tolerance bands (wider at offsets, +/-0.727 s, than onsets, +/-0.5 s). The per-annotator annotations and analysis code are available at   \url{https://github.com/WSCSports/MTLLFM-temporal-laughter-localization}.
\end{abstract}

\section{Introduction}\label{sec:intro}

\begin{figure}[tbp]
\centering
\includegraphics[width=1\columnwidth]{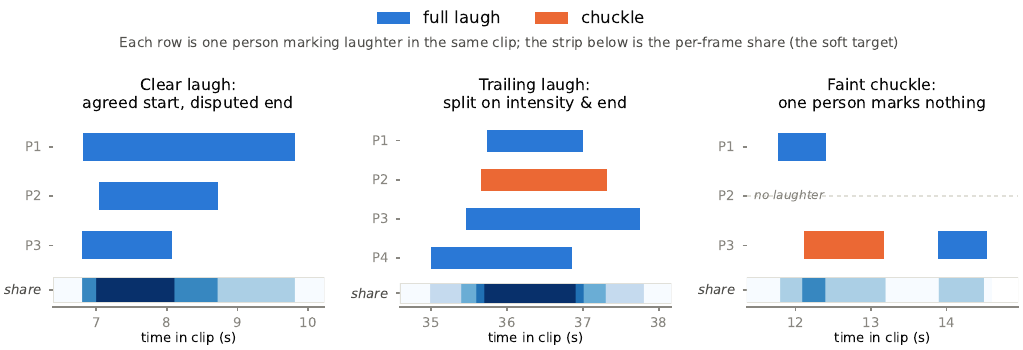}
\caption{Laughter has no single boundary. Three clips from the corpus, each with a disputed
extent: a clear laugh whose end annotators place differently; a trailing laugh split on both
intensity (full vs.\ chuckle) and offset; and a faint chuckle one annotator does not mark at
all. Evaluating against a single ground truth hides this structured disagreement.}\label{fig:concept}
\end{figure}

When does a laugh begin - and, harder, when does it end? Ask several trained annotators to
mark the same clip and they return different answers (Fig.~\ref{fig:concept}): aligned on the
sharp attack of a clear laugh, scattered across its gradual decay, and split on whether a
faint chuckle is a laughter event at all. This is not an exotic corner case. Laughter is among
the most frequent and most studied affective signals in real-world video, and localizing it in
time, not merely detecting its presence, underpins applications from affective indexing to
highlight generation. Yet laughter localization benchmarks, like temporal affect benchmarks
generally~\cite{li2022tel,zhang2023cvpr,kollias2022abaw}, evaluate systems against a
\emph{single} reference annotation (henceforth \emph{single-GT}), treating every difference
between annotators as noise around one true boundary.

This paper argues, with data, that these differences are not noise. Our thesis:
\emph{single-ground-truth evaluation misrepresents temporal laughter localization. Laughter
boundaries are not point events; annotators disagree about them in structured, measurable
ways; and that disagreement should be modeled rather than averaged away.} To test it, we
re-annotate the SMILE-Temporal benchmark~\cite{hyun2024smile} with multiple independent
annotators per video and release \textbf{SMILE-Disagreement}, to our knowledge the first
publicly available, per-event, multi-annotator temporal-boundary corpus for laughter in
affect video. We state this claim narrowly and position it against LEAD~\cite{imoto2025lead}, Vargas-Quiros~\cite{vargasquiros2023laughter}, and
GEBD~\cite{shou2021gebd} in Sec.~\ref{sec:related}.

The disagreement we find is not random. Annotators agree well overall (Krippendorff's
$\alpha = \frameAlpha$~\cite{krippendorff2011alpha} at 100\,ms, stable across bin sizes and annotator counts), so the
signal is trustworthy; but \emph{where} they disagree is systematic. Two regularities stand
out. First, offsets are $\asymRatioLow$--$\asymRatioHigh\times$ noisier than onsets: a laugh has a sharp attack
but a gradual decay, so its end is genuinely harder to pin down than its start. Second,
low-intensity ``chuckles'' are contested far more often than full laughs
($\chuckleDisagree\%$ vs.\ $\fullDisagree\%$ of events), an intensity effect that survives
every control we apply: modality, duration, annotator count, choice of event matcher, and the
(imperfect) reliability of the intensity label itself. This is not only about whether a laugh
is marked at all: even when all annotators agree a chuckle occurred, its boundaries are about
twice as spread out as a full laugh's at the same duration. Disagreement is, moreover,
predictable from event attributes alone (Sec.~\ref{sec:findings:chuckle}).
In short, the ambiguity is a property of the events, not annotator caprice, and averaging it
away is not a harmless convenience but a modeling error.

The practical casualty is evaluation. Because annotators genuinely differ, any single one is
an arbitrary reference: the same system's frame-F1 moves by \singleGtSwing\ on average
depending on which annotator is chosen as ground truth, and - in controlled experiments
with synthetic systems of known quality - single-GT ranks the genuinely
better system above the worse one only $\sgRank\%$ of the time, versus ${\sim}80\%$ when scoring
against \emph{all} annotators (Sec.~\ref{sec:eval}). We therefore propose and validate a
\emph{disagreement-calibrated} protocol: score a prediction against the whole annotator
distribution, with a per-boundary tolerance band conformally calibrated to how much humans
themselves disagree, wider at offsets than at onsets.

\noindent In summary, our contributions are:
\begin{enumerate}
  \item \textbf{Dataset.} SMILE-Disagreement: \nBenchmark\ videos with 3--5 independent
        annotators each, carrying \nMarksBench\ annotator marks (\nMatchedEvents\ distinct
        laughter events after cross-annotator matching) with intensity and source labels,
        released as per-annotator distributions with all analysis code
        (Sec.~\ref{sec:dataset}).
  \item \textbf{Findings.} A perspectivist analysis that ports the
        learning-with-disagreement lens from categorical NLP to temporal boundaries and
        shows the disagreement is structured: intensity-linked, offset-concentrated, robust
        to five artifact controls, and predictable from event attributes
        (Secs.~\ref{sec:related}, \ref{sec:findings}).
  \item \textbf{Evaluation.} A disagreement-calibrated protocol (a soft distributional
        target plus a conformal human-spread band; specified in Sec.~\ref{sec:eval:protocol}),
        validated to remove the reference-annotator arbitrariness that makes single-GT
        rankings unreliable, and exercised on a real public laughter localizer - whose
        errors concentrate exactly where human disagreement does (Sec.~\ref{sec:eval}).
\end{enumerate}

\section{Related Work}\label{sec:related}

\textbf{Human label variation and perspectivism.} A decade of work has challenged the
assumption that annotation converges on one correct label: disagreement reflects genuine
properties of the data rather than annotator error~\cite{aroyo2015truth}, and is
reproducible and inherent rather than resolvable by more careful
annotation~\cite{pavlick2019inherent}. The resulting perspectivist,
learning-with-disagreement program~\cite{uma2021survey,plank2022disagreement} keeps the
variation and evaluates against it - via the soft label distributions of the LeWiDi
benchmark~\cite{leonardelli2023lewidi}, label-distribution learning~\cite{geng2016ldl},
and, in vision, full human label distributions over CIFAR-10~\cite{peterson2019human}. All
of this concerns \emph{categorical} labels, where disagreement is a rate
over classes. We port the perspectivist question to \emph{temporal boundaries} in affect
video, where a label is an interval and disagreement acquires geometry: it can concentrate
at onsets or offsets, scale with event intensity, and widen or narrow per event.

\textbf{Temporal boundary ambiguity in video.} Vision has long had evidence that temporal
boundaries are soft: recognition accuracy depends on which temporal extent of an action one
assumes~\cite{satkin2010temporal}, conscientious workers disagree substantially on temporal
extents~\cite{sigurdsson2016muchado} and interaction bounds~\cite{moltisanti2017trespassing},
and diagnostic analyses identify imprecise boundary localization as a principal error
source in temporal action detectors~\cite{alwassel2018detad}.
Detection~\cite{shou2021gebd} responds by collecting five annotators per video and
evaluating against each separately. These efforts, however, treat the variation as noise to
be managed - averaged away, regularized against, or (in GEBD~\cite{shou2021gebd}) attached to
\emph{instantaneous} cut points. Affective events extend over time, with an onset and an
offset whose uncertainties differ - the systematic asymmetry we quantify in
Sec.~\ref{sec:findings:asymmetry} has no analogue for point events - and we release the
variation itself, per event and per annotator, as the object of study.

\textbf{Multi-annotator affect, laughter, and sound-event corpora.} Closest to our
resource, the LEAD dataset~\cite{imoto2025lead} releases strong labels from 20 annotators
for general sound events and analyzes how onset/offset labels vary, but stops short of a
disagreement-aware evaluation. For laughter specifically, Vargas-Quiros et
al.~\cite{vargasquiros2023laughter} compare annotation modalities with two annotators per
segment and aggregate their labels rather than releasing the per-event boundaries of
${\geq}3$ distinct annotators.
Affect corpora (RECOLA~\cite{ringeval2013recola}, SEWA~\cite{kossaifi2019sewa}) ship
multiple raters as dimensional traces rather than discrete event boundaries, and the ABAW
competition series~\cite{kollias2022abaw} evaluates affect recognition against a single
reference annotation per video. The classic laughter resources
(MAHNOB~\cite{petridis2013mahnob}, AMI~\cite{carletta2005ami}) and the detection and
segmentation line built on
them~\cite{truong2007laughter,petridis2008fusion,escalera2009multimodal,gillick2021laughter,omine2024laughterseg}
all treat a laugh's extent as given - as do the SMILE benchmark~\cite{hyun2024smile} we
re-annotate, the multimodal localizer trained against it~\cite{hanania2026mtllfm}, and its
successor SMILE-Next~\cite{lee2026smilenext}, which scales laughter understanding to
language models but neither localizes boundaries in time nor models annotator
disagreement.
Our corpus occupies the intersection these leave open
(Table~\ref{tab:diff}): per-event onset/offset \emph{distributions} from ${\geq}3$ distinct
annotators, on laughter/affect video, publicly released.

\section{The SMILE-Disagreement Dataset}
\label{sec:dataset}

We re-annotate the videos of SMILE-Temporal~\cite{hyun2024smile}, a temporal
laughter-localization benchmark, with multiple independent annotators, each marking laughter
onset and offset boundaries. To our knowledge it is the first publicly released, per-event, multi-annotator temporal-boundary corpus for laughter in
affect video; Table~\ref{tab:diff} positions it against the closest prior resources.

\subsection{Annotation protocol}
\label{sec:dataset:collection}
Six trained annotators labeled the videos independently, without seeing one another's labels.
For each perceived laughter event an annotator marked its onset and offset and
assigned three attributes: \textbf{intensity} (full laughter vs.\ chuckle);
\textbf{source}, the dominant cue the laughter was perceived through
(acoustic, visual, or both); and \textbf{speaker} (audience,
\nEvAud\ of the \nMarksBench\ benchmark marks, vs.\ individual speaker, \nEvSpk).
Source records the dominant channel, not an exclusive one: a visual cue may still contribute
where a laugh is labelled acoustic, possibly without the annotator consciously registering
it, so the acoustic-dominant split (Sec.~\ref{sec:dataset:composition}) reflects the
primary perceived channel, not the absence of visual information.

The benchmark is every video with ${\geq}3$ distinct annotators (\nBenchmark\ videos;
per-video counts in Table~\ref{tab:raters}), the accepted minimum for reliability analysis;
the count is uneven, and Sec.~\ref{sec:findings} confirms every finding holds when videos are
subsampled to three. It carries \nMarksBench\ annotator marks (one region drawn by one
annotator; the full campaign collected \nEvents, the rest repeats or below-threshold videos),
which cross-annotator matching merges into \nMatchedEvents\ distinct events.

\begin{table}[t]
\centering
\small
\begin{tabular}{lccc}
\toprule
Distinct annotators per video & 3 & 4 & 5 \\
\midrule
\# videos & \nRaterThree & \nRaterFour & \nRaterFive \\
\bottomrule
\end{tabular}
\caption{Per-video annotator counts in the \nBenchmark-video benchmark (${\geq}3$ distinct
annotators; same-annotator repeats removed).}
\label{tab:raters}
\end{table}

\subsection{Annotator consistency}
\label{sec:dataset:annotators}
A per-annotator profile (anonymized A1-A6, matching the released data) is given in the
supplementary. No annotator is an outlier: frame-level agreement with the majority of the
others spans \annFOneLow-\annFOneHigh\ F1, productivity and existence rates are comparable,
and onset/offset style biases stay within a few tenths of a second. The disagreement
structure analyzed in Sec.~\ref{sec:findings} is therefore a property of the videos themselves,
not of one annotator's labeling style.

\subsection{Composition}
\label{sec:dataset:composition}
Of the 672 benchmark videos, 578 contain at least one annotated laughter event; these 578 total 4.49 hours of video (median 26.0 s, IQR 15.9–35.6 s),
and after matching
holds \nMatchedEvents\ distinct events - \evPerVideoMean\ per video and \evPerMin\ per
annotated minute - with laughter marked by at least one annotator covering \laughShareUnion\%
of recorded time (\laughShareMaj\% by majority vote). Median event duration is \durFullMed\,s
for full laughter vs.\ \durChuckleMed\,s for chuckles, one reason low-intensity events are
easier to miss outright (Sec.~\ref{sec:findings:chuckle}). Across all \nEvents\ campaign marks,
\pctFullIntensity\% are full laughter and \pctChuckleIntensity\% chuckles. By cue, annotators
judged \pctAcoustic\% acoustic-dominant, \pctBoth\% audiovisual, and \pctVisual\% visual-only;
laughter's most salient cue is usually its sound, so an acoustic-leaning split is expected, and
this is the dominant channel rather than the only one, on what remains full audiovisual video
(Sec.~\ref{sec:dataset:collection}). The audiovisual minority is where the difficulty
concentrates: visually-involved events are disagreed about \emph{more} than acoustic-only ones
($\disVisual\%$ vs.\ $\disAcousticOnly\%$; 95\% Wilson CI $[\disVisualLow, \disVisualHigh]$,
$n{=}\nVisualClusters$ matched events), so the corpus still exercises the subtle, cross-modal
cases an audiovisual system must resolve. Finally, $\pctAudienceEv\%$ of marks are audience
laughter: the source videos are edited, broadcast-style content, where laughter comes
predominantly from an audience rather than a single on-screen speaker. This is a domain
property, not a confound behind our findings: the offset asymmetry replicates within both the
audience and the individual-speaker subsets (Sec.~\ref{sec:findings:asymmetry}).

We publicly release SMILE-Disagreement - the per-annotator onset/offset
annotations and metadata as a JSON file, for
non-commercial research purposes, together with the full analysis code, at
\url{https://github.com/WSCSports/MTLLFM-temporal-laughter-localization}. The underlying video content is part
of the original SMILE dataset~\cite{hyun2024smile} and is not redistributed; our
release adds only the multi-annotator annotation layer.

\subsection{Relation to prior multi-annotator resources}
\label{sec:dataset:diff}
Multi-annotator boundary annotation exists in adjacent settings, and we claim no priority over
it; Table~\ref{tab:diff} positions our corpus against the closest work. Each prior resource
lacks one of our four defining properties: LEAD~\cite{imoto2025lead} annotates general
sound events, not affect video; Vargas-Quiros et al.~\cite{vargasquiros2023laughter} use two
annotators per segment (48 total), aggregated and unreleased; GEBD~\cite{shou2021gebd} marks
instantaneous cut points rather than onset/offset intervals; and
RECOLA~\cite{ringeval2013recola} / SEWA~\cite{kossaifi2019sewa} provide continuous dimensional
traces rather than discrete per-event boundaries. Our contribution is the \emph{intersection}
they leave open: publicly released, per-event onset/offset \emph{distributions} from
${\geq}3$ distinct annotators on laughter/affect video.

\begin{table*}[t]
\centering
\small
\setlength{\tabcolsep}{4pt}
\resizebox{\textwidth}{!}{%
\begin{tabular}{lccccc}
\toprule
Resource & Modality & \# Annotators/item & Per-event onset/offset & Released as distribution & Domain \\
\midrule
\textbf{Ours (SMILE-Disagreement)} & AV (acoustic-dom.) & 3-5 distinct & \cmark & \cmark & laughter / affect video \\
LEAD~\cite{imoto2025lead}          & audio & 20 & \cmark & \cmark & general sound events \\
Vargas-Quiros~\cite{vargasquiros2023laughter} & AV & 2/seg.\ (48 tot.) & \cmark & \xmark\ (not released) & laughter \\
GEBD~\cite{shou2021gebd}           & video & 5 & cut-points (instant.) & partial & generic boundaries \\
RECOLA / SEWA~\cite{ringeval2013recola,kossaifi2019sewa} & AV & multiple & continuous traces & \cmark & dimensional affect \\
\bottomrule
\end{tabular}%
}
\caption{Positioning against the closest multi-annotator resources. Our scoped novelty is
the combination of per-event onset/offset distributions, public release,
${\geq}3$ distinct annotators, and the laughter/affect-video domain: no
prior row occupies all four columns at once.}
\label{tab:diff}
\end{table*}

\section{Analysis and Findings}
\label{sec:findings}

We treat the multi-annotator labels not as noise to be averaged into a single ground truth
but as a \emph{distribution} to be characterized; all statistics are computed on the
\nBenchmark-video benchmark of Sec.~\ref{sec:dataset} (a summary table is in the
supplementary). We first establish that the agreement signal is stable - the claim that
disagreement is \emph{structured} is only meaningful if it is - and then show that where
annotators disagree is governed by two factors: \emph{where} a boundary lies (onset vs.\
offset) and \emph{how intense} the laughter is.

\subsection{Annotators agree well, and robustly}
\label{sec:findings:agreement}
Discretizing each video into 100\,ms frames and treating laughter presence as a binary label,
frame-level inter-annotator reliability is Krippendorff's
$\alpha = \frameAlpha$.\footnote{Computed with the standard \texttt{krippendorff} library
(nominal level, missing data for annotators who did not label a given video); an independent
re-implementation agrees to four decimals.} Because frames within a video are strongly
autocorrelated, we report a video-level (clustered) bootstrap rather than treating frames as
independent units: $\alpha \in [\bootAlphaLow, \bootAlphaHigh]$ (95\% CI, $B{=}1000$, videos
resampled with replacement). The value is well above the chance floor and not an artifact of
our choices - it holds across frame sizes from 40 to 500\,ms
($\alpha \in [\alphaBinLow, \alphaBinHigh]$) and is unchanged when every video is subsampled
to exactly three annotators ($\alpha = \frameAlphaSub$), so the uneven annotator count is not
inflating it. Agreement on the \emph{existence} of laughter is likewise high but not total:
annotators agree on whether laughter is present in $\existUnanimous\%$ of videos, leaving
$\existContested\%$ contested. The signal is real; the structure in the disagreement, below,
is therefore interpretable.

\subsection{Offsets are fuzzier than onsets}
\label{sec:findings:asymmetry}
\begin{figure}[tbp]
\centering
\begin{subfigure}{0.49\columnwidth}
  \centering
  \includegraphics[width=\linewidth]{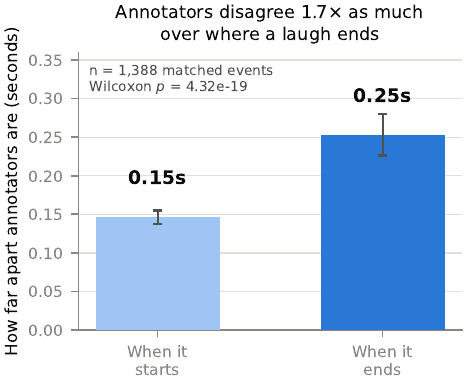}
  \caption{}\label{fig:asymmetry}
\end{subfigure}\hfill
\begin{subfigure}{0.49\columnwidth}
  \centering
  \includegraphics[width=\linewidth]{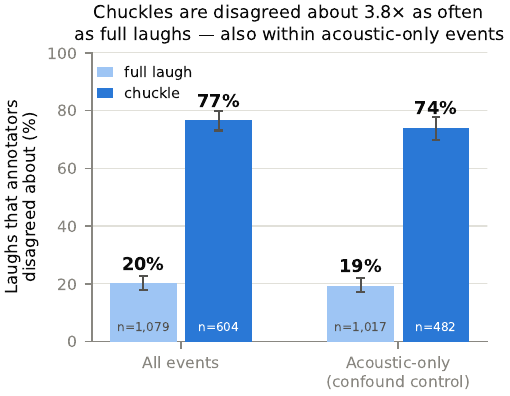}
  \caption{}\label{fig:chuckle}
\end{subfigure}
\caption{The two headline regularities. \textbf{(a)} Annotators are about
$\asymRatio\times$ as far apart on where a laugh \emph{ends} as on where it \emph{begins}:
mean across-annotator standard deviation over commonly-marked events, 95\% CIs (one-sided
Wilcoxon signed-rank; greedy matcher - ratio $\asymRatioLow$--$\asymRatioHigh\times$
across matchers). \textbf{(b)} Disagreement rate by intensity (bars: rate with 95\% Wilson
CI, $n$ inside each bar): a chuckle is disagreed about $\chuckleDisagree\%$ of the time
versus $\fullDisagree\%$ for a full laugh; the right group repeats within acoustic-only
events, so the gap is driven by \emph{intensity}, not modality.}
\label{fig:findings}
\end{figure}

For events that multiple annotators localized in common, we measure the across-annotator
standard deviation of each boundary. Offsets are markedly noisier (Fig.~\ref{fig:asymmetry}):
the mean offset spread ($\offsetSD$\,s) is about $\asymRatio\times$ the onset spread
($\onsetSD$\,s;
one-sided Wilcoxon signed-rank $p = \asymWilcoxonP$, video-level bootstrap 95\% CI on the
greedy-matcher ratio $[\bootAsymLow, \bootAsymHigh]$).
It is not an artifact of the
corpus's dominant communal audience laughter, whose decay many people could stop marking at
different times: it holds \emph{within} audience-labeled events ($\asymRatioAud\times$,
bootstrap 95\% CI $[\asymRatioAudLow, \asymRatioAudHigh]$, $n{=}\nAsymAud$) and, indicatively,
within the small individual-speaker group ($\asymRatioSpk\times$, CI
$[\asymRatioSpkLow, \asymRatioSpkHigh]$, $n{=}\nAsymSpk$) - single-person laughs show the
same offset fuzziness.
The cause is acoustic. Measuring the RMS energy envelope around each co-marked event (half-max
crossing times in a padded window, so they do not depend on the human boundaries), the central
peak rises faster than it falls: the decay is $\decayOverAttack\times$ the attack (median;
video-level bootstrap 95\% CI $[\decayOverAttackLow, \decayOverAttackHigh]$; decay exceeds
attack in $\fracDecayGtAttack\%$ of $\nEnvEvents$ events). A sharp attack but gradual decay
makes the endpoint acoustically the less determined boundary - the direction of the
annotation asymmetry. The effect is aggregate, not an event-wise dose--response (within
events, decay length does not predict offset spread, Spearman $\rho{=}\rhoDecayOffset$), so we
read the offset fuzziness as a general perceptual consequence of the gradual decay. Any
evaluation should therefore grant more tolerance at offsets than at onsets, a prescription we
make precise in Sec.~\ref{sec:eval}.

\subsection{Intensity drives disagreement}
\label{sec:findings:chuckle}
The second and larger factor is intensity, and it acts on \emph{both} whether an event is
marked and where its boundaries fall.

\noindent\textbf{Existence.} Calling an event \emph{disagreed} when not all of a video's
annotators marked it (greedy-overlap matcher; an event's intensity is the majority vote of its
co-markers' labels), a chuckle is disagreed about $\chuckleDisagree\%$ of the time versus only
$\fullDisagree\%$ for full laughter (Fig.~\ref{fig:chuckle}; video-level clustered bootstrap
95\% CI on the gap $[\bootGapLow, \bootGapHigh]$ points) - equivalently, $\chuckleOfContested\%$
of all disagreed events are chuckles, against just $\chuckleOfUnanimous\%$ of unanimous ones.
Low-intensity laughter is where humans disagree; unambiguous full laughter is where they
agree.

Five measurement artifacts could mimic this, and each fails to explain it (rates also in the
supplementary summary table). \emph{Modality:} restricting to acoustic-only events leaves the gap
essentially unchanged. \emph{Duration:} chuckles are shorter and easier to miss outright, yet
the gap holds at \chuckleGapStratMin--\chuckleGapStratMax\ points \emph{within every duration
quartile}, and a logistic model with log-duration, position, and modality as covariates still
assigns a chuckle $\chuckleOddsRatio\times$ the odds of disagreement. \emph{Annotator count:}
subsampling every video to exactly three annotators leaves the rates essentially unchanged
($\chuckleDisagreeSub\%$ vs.\ $\fullDisagreeSub\%$). \emph{Event matcher:} IoU and midpoint
matching give \chuckleMatchLow--\chuckleMatchHigh\% vs.\ \fullMatchLow--\fullMatchHigh\%.
\emph{The intensity label itself:} co-markers disagree on it for $\mixedIntPct\%$ of
jointly-marked events ($\alpha = \intAlpha$), consistent with intensity being graded rather
than binary - but such shared label noise attenuates rather than inflates the gap, and
restricting to intensity-unanimous events in fact widens it ($\chuckleDisagreeUnanInt\%$ vs.\
$\fullDisagreeUnanInt\%$). Nor is the structure specific to the dominant audience laughter:
within audience-labeled events the rates are $\chuckleDisagreeAud\%$ vs.\ $\fullDisagreeAud\%$,
and the small individual-speaker group points the same way ($\chuckleDisagreeSpk\%$ vs.\
$\fullDisagreeSpk\%$; $n{=}\nChSpk$/$\nFullSpk$, indicative only).

Disagreement is thus substantially \emph{predictable} from event attributes: the same logistic
model reaches AUC $\disagreeAUC$ under 5-fold cross-validation grouped by video
($\disagreeAUCNoInt$ without the intensity feature). Because intensity is itself an annotation
rather than an acoustic measurement, we bound the resulting circularity against the signal. On
all $\nAudioVideos$ benchmark videos (the audio is fully recoverable), a label-free acoustic
loudness proxy (mean RMS, duration-independent) already separates full laughs from chuckles
(AUC $\proxyAUCsep$; full laughs are about twice as loud), and substituting it for the
perceptual label lifts the disagreement model from $\proxyAUCnoint$ (no intensity signal at
all) to $\proxyAUCrms$ - below the $\proxyAUCperc$ the perceptual label reaches, but clearly
above the no-intensity baseline. The intensity effect therefore has a genuine acoustic basis
and is not merely an artifact of the pool's own labels.

\noindent\textbf{Boundaries.} Intensity structures not only existence but boundary spread,
even on events multiple annotators agree exist. On co-marked events (greedy matcher, same
spread statistic as Sec.~\ref{sec:findings:asymmetry}), chuckles are fuzzier than full laughs
(onset $\chOnsetSD$ vs.\ $\fullOnsetSD$\,s, Mann--Whitney $p = \bndOnP$; offset $\chOffsetSD$
vs.\ $\fullOffsetSD$\,s, $p = \bndOffP$; $n = \nChCom$ vs.\ $\nFullCom$), and the pooled means
\emph{understate} this for the same reason as above - shorter events have tighter absolute
spreads - so within duration quartiles the chuckle offset spread is
$\bndQuartRatioLow$--$\bndQuartRatioHigh\times$ the full-laugh spread, in all four. The
comparison is conservative: co-marked chuckles are by construction the \emph{least} ambiguous
chuckles. The offset--onset asymmetry, in turn, holds \emph{within both} intensities
($\chAsymRatio\times$ for chuckles, $\fullAsymRatio\times$ for full laughs), so the two axes
are separable: intensity makes an event both likelier to be missed entirely and, when found,
roughly twice as uncertain at the boundaries, while the offset concentration is a property of
laughter's temporal envelope regardless of intensity.
\section{Disagreement-Calibrated Evaluation}
\label{sec:eval}

The findings of Sec.~\ref{sec:findings} imply that a single annotator is a poor stand-in for
``ground truth.'' This section makes the damage concrete, then proposes and validates an
evaluation that scores a system against the \emph{distribution} of human labels. The
protocol is built from the human annotations alone - no trained model is required to
construct or validate it (Sec.~\ref{sec:eval:real} then applies it to one).

\subsection{A single ground truth is arbitrary}
\label{sec:eval:arbitrary}
We first quantify the arbitrariness. Treat each annotator as a ``system'' and score it, at
the frame level (100\,ms), against each of the other annotators, one at a time. Two facts
emerge.
First, human--human agreement \emph{averages}
a frame-F1 of about \humanCeiling: this, not $1.0$, is the realistic ceiling for any system,
yet a single-ground-truth (single-GT) benchmark implicitly treats perfection as attainable. Second, and more
damaging, a system's measured score depends heavily on \emph{which} annotator is chosen as
reference: for each (video, system) pair we take the max--min range of its frame-F1 across
the available reference choices; the mean range is \singleGtSwing\ F1
(Fig.~\ref{fig:ranking}a).
Under single-GT, the reported number is partly an artifact of an arbitrary choice.

\subsection{Scoring against the human distribution}
\label{sec:eval:protocol}
We instead score a prediction against the annotator distribution, with two ingredients.

\emph{(i) A soft target.} Let $y^a_i \in \{0,1\}$ be annotator $a$'s laughter label on
100\,ms frame $i$, and let $s_i = \frac{1}{k}\sum_{a=1}^{k} y^a_i$ be the per-frame fraction
of the $k$ annotators marking laughter. A binary prediction $\hat{y}$ is scored by the fuzzy
Jaccard overlap (soft-IoU)
\begin{equation}
\mathrm{sIoU}(\hat{y}, s) \;=\; \frac{\sum_i \min(\hat{y}_i, s_i)}{\sum_i \max(\hat{y}_i, s_i)},
\label{eq:siou}
\end{equation}
so a prediction earns graded credit for partially-agreed events instead of being right or
wrong against one person. When the denominator vanishes (a video with no marked
laughter and no prediction) we score $\mathrm{sIoU}{=}1$: correct silence.

\emph{(ii) A conformally calibrated tolerance band.} For each boundary type
$b \in \{\text{onset}, \text{offset}\}$, we hold out one annotator at a time and compute the
nonconformity score $r = |t - m|$, where $t$ is the held-out annotator's boundary and $m$ is
the median of the matched boundaries of the remaining annotators (events matched by temporal
overlap; only events with ${\geq}2$ co-markers, annotators who also marked the event,
contribute). We pool the $r$'s of type $b$
across all events and held-out annotators ($n{=}\nBandCal$ per type) and take the
conformal quantile $q_b = r_{(\lceil (n+1)(1-\alpha) \rceil)}$, the
$\lceil (n+1)(1-\alpha) \rceil$-th smallest score: the half-width calibrated to cover a
$1{-}\alpha$ fraction of held-out \emph{human} boundaries. At
$\alpha{=}0.1$ the band is $\pm\bandOnsetNinety$\,s at onsets but $\pm\bandOffsetNinety$\,s
at offsets (\bandRatioNinety$\times$ wider): the metric is most lenient where humans
themselves are least certain, inheriting by construction the boundary asymmetry of
Sec.~\ref{sec:findings:asymmetry} rather than ignoring it.

These global widths are
driven by a heavy tail of ambiguous events; an \emph{adaptive} variant, which rescales
each $r$ by the local spread of the co-markers' boundaries, attains the same coverage
with narrower per-event widths, widening only on the events where the annotators
themselves disagree. Its specification, and two further calibration details, are in the
supplementary.

A prediction's \emph{band score} is the fraction of its matched boundaries falling within
$\pm q_b$ of the corresponding human median. Predicted events that overlap no human event
contribute no boundaries to it: hallucinations and misses are punished by the soft-IoU
term instead - a hallucinated event's frames enter the denominator of
Eq.~(\ref{eq:siou}) at full weight, a missed event's frames enter it at their human
weight $s_i$, and neither adds to the numerator.
Sec.~\ref{sec:eval:validation} probes this division of labor for exploitability.

Two caveats delimit what the calibration promises. First, coverage is \emph{marginal} (it
holds on average over the pooled residuals, not per video or annotator), and pooling
treats residuals as exchangeable although they cluster within videos, within annotators,
and within events (each event yields several leave-one-out residuals). The annotator axis
we verify empirically: calibrating on
five annotators and testing on the sixth yields $\covOutLow$--$\covOutHigh$ (mean
$\covOutMean$, target $0.90$) across all six annotators and both boundary types -
approximately calibrated, with mild under-coverage at the extremes. That check pools
events, so the other clustering axes remain assumptions, not verified properties.
For a system's predictions the band is a
\emph{human-calibrated tolerance}, \emph{not} a coverage guarantee: it asks whether a
prediction deviates from the consensus by more than one human deviates from the rest. A
model's residuals need not be exchangeable with human ones, and Sec.~\ref{sec:eval:real}
shows they are not.

\emph{Composition.} The full protocol score is the unweighted mean of the frame-level
soft-IoU~(\ref{eq:siou}) and the event-level band score; both are computed per video and
macro-averaged (unweighted mean over videos), so long videos do not dominate.
Sec.~\ref{sec:eval:validation} validates each ingredient separately, the composition, and
its insensitivity to the mixing weight.
The supplementary illustrates the full protocol on a real event;
Sec.~\ref{sec:related} situates it as, to our knowledge, the first temporal-boundary
evaluation with \emph{human-calibrated} tolerance (GEBD~\cite{shou2021gebd}, the nearest
precedent, scores instantaneous cut points against each annotator separately under a
convention-chosen tolerance).


\subsection{Validation: does it rank systems better?}
\label{sec:eval:validation}
\begin{figure}[tbp]
\centering
\includegraphics[width=0.75\columnwidth]{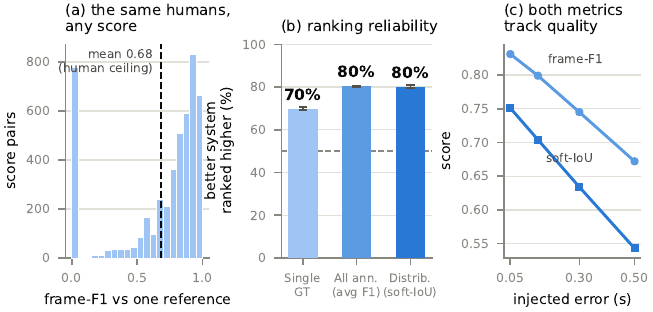}
\caption{Single-GT evaluation and its repair. \textbf{(a)} The cause: scoring one
annotator-as-system against each single reference spreads scores across the entire
$[0,1]$ range - the reported number depends on who grades. \textbf{(b)} The damage:
single-GT ranks the better synthetic system above a one-tier-worse one far less reliably
than either rule that uses all annotators (bars: mean; whiskers: min--max over five
seeds). \textbf{(c)} The validity check: both metrics decrease monotonically with
injected boundary error; only ranking \emph{reliability} differs.}
\label{fig:ranking}
\end{figure}

A good evaluation should rank a better system above a worse one, so we test exactly that.
We inject synthetic ``systems'' of \emph{known} quality, the consensus localization
perturbed with increasing boundary jitter (four tiers, so tier order is ground-truth
quality order), and ask, for each adjacent pair and over five random seeds, whether a
scoring rule ranks the better system higher (Fig.~\ref{fig:ranking}). Three rules are
compared, differing in how they use the annotators: frame-F1 against a \emph{single}
arbitrary annotator; frame-F1 \emph{averaged over all} annotators; and soft-IoU against the
annotator \emph{distribution}.

Single-GT gets the ranking right only $\sgRank\%$ of the time; both distribution-aware rules
reach ${\sim}80\%$ (multi-annotator average \multiRank\%, soft-IoU \calRank\%). The
soft-IoU--vs--single-GT gap carries a paired-bootstrap 95\% CI of
\rankGapLow--\rankGapHigh\ points; that bootstrap resamples pooled ranking trials
independently, ignoring within-video correlation, so the true interval can only be wider.
The gain decomposes cleanly: essentially all of it comes from
evaluating against \emph{all} annotators rather than an arbitrary one (the choice
single-GT benchmarks silently make), while the soft-IoU rule matches, but does not
exceed, the averaged rule's reliability on these synthetic systems.
What the distributional form adds is therefore not ranking power: the reference-annotator
degree of freedom is already removed by \emph{any} rule that uses all annotators, so the
form's own contribution is graded credit for partially-agreed events and the calibrated
tolerance band of Sec.~\ref{sec:eval:protocol}, whose role the next paragraph tests. The
decomposition is also robust to how the synthetic systems are generated: perturbing a
random annotator's labels instead of the consensus gives
$\sgRankAnnOrigin\%$ / $\multiRankAnnOrigin\%$ / $\calRankAnnOrigin\%$.

The tolerance band plays a deliberately different role. Scored alone it declares
$\bandTiePct\%$ of adjacent-tier pairs \emph{tied}: both systems sit within the
human-calibrated tolerance, and an evaluation aligned with human ambiguity should indeed
refuse to rank differences smaller than human disagreement. When it does decide, it
decides correctly $\bandDecidedAcc\%$ of the time; and the full protocol (soft-IoU +
band) preserves ranking reliability ($\fullRank\%$; flat at
\fullRankWLow--\fullRankWHigh\% across mixing weights of $0.25$--$0.75$, so the unweighted
mean is a convention, not a tuned parameter). Nor is the composition gamed by the obvious
boundary-only play: a degenerate ``system'' predicting one well-placed $0.2$\,s sliver per
video scores \degFull\ under the full protocol versus \worstTierFull\ for even the worst
jittered tier - its soft-IoU collapses, and even its band score is only \degBand.

The same test also bounds how many annotators a temporal benchmark needs. When the soft
target is restricted to $k$ subsampled annotators on the
${\geq}4$-annotator videos (curve in the supplementary), ranking reliability climbs from
$\annKOneRank\%$ ($k{=}1$, the single-GT regime on this subset) through $\annKTwoRank\%$
($k{=}2$) and $\annKThreeRank\%$ ($k{=}3$) toward the all-annotator $\annAllRank\%$ on the
same subset: most of the benefit comes from the second and third annotator.

\subsection{On a real localizer}
\label{sec:eval:real}
Synthetic tiers validate a metric; a real system is what the protocol is \emph{for}. No
laughter-specific audiovisual localizer is publicly released: UnAV-100
localizers~\cite{geng2023unav} carry a laughter class among their hundred categories but
ship no laughter-specialized checkpoint, and this corpus is precisely the resource such a
model would need. We therefore run the strongest public trained laughter localizer
available, the audio-only segmentation model of Omine et
al.~\cite{omine2024laughterseg}, with its released checkpoint and inference settings,
over the \blN\ benchmark videos with at least one annotated laugh, yielding
\blNEvents\ predicted events. (On the \blNegVids\ additional videos where \emph{no}
annotator marked laughter it predicts only \blNegEvents\ events, so it almost never
hallucinates.) An audio-only system is a reasonable first entry for this corpus, whose
reported cue is \pctAcoustic\% acoustic (Sec.~\ref{sec:dataset}); its errors can be read
against the annotators' own cue labels.

\begin{table}[tbp]
\centering
\caption{The protocol on the real localizer ($n{=}\blNMatched$ matched events). The
\emph{Human} column is the matching reference per row: the frame-F1 ceiling and the
annotator-as-system swing (row~1); the band-coverage target ($0.90$ by construction; the
empirical annotator-out range \covOutLow--\covOutHigh\ is measured under the \emph{global}
band); and tier ranking on synthetic tiers. Coverage is the within-band fraction per
boundary type (onset/offset); the scalar band score pools both. Tier-ranking values are
single-GT\,/\,multi\,/\,calibrated\,/\,full.}
\label{tab:realmodel}
\footnotesize
\renewcommand{\arraystretch}{1.05}
\setlength{\tabcolsep}{5pt}
\begin{tabular}{lcc}
\toprule
Metric & Model & Human \\
\midrule
Single-GT frame-F1 (swing/vid.)  & \blSG\ (\blSwing)                       & \humanCeiling\ (\singleGtSwing) \\
Multi-annotator frame-F1         & \blMulti                                & --- \\
Soft-IoU / band score            & \blCal\ / \blBand                       & --- \\
Coverage global, ons./offs.      & \blCovOnG\ / \blCovOffG                 & $0.90$ \\
\quad adaptive, ons./offs.       & \blCovOnA\ / \blCovOffA                 & \covOutLow--\covOutHigh \\
Miss rate, contested / unanim.   & \blMissContested\% / \blMissUnanimous\% & --- \\
\quad chuckle / full             & \blMissChuckle\% / \blMissFull\%        & --- \\
Tier rank, SG/mul/cal/full (\%)  & \blRankSG/\blRankMulti/\blRankCal/\blRankFull & \sgRank/\multiRank/\calRank/\fullRank \\
\bottomrule
\end{tabular}
\end{table}

The effects the protocol was built for all appear on the real system
(Table~\ref{tab:realmodel}). \textbf{The swing is real:} the reference-annotator choice
moves the model's per-video frame-F1 by even more than the swing measured with
annotators-as-systems, though a system operating farther below the ceiling has more room
to swing, so we read the direction, not the magnitude.
\textbf{The band does metric work, most clearly at offsets.} Under the global band the
model falls short of the $0.90$ human construction target at both boundary types; under
the adaptive band its onset coverage sits \emph{within} the human annotator-out range
(itself measured under the global band, so we read the adaptive comparison as indicative
only), and the offset shortfall persists. The $0.90$ is a property of the human pool, not
a guarantee the model inherits (Sec.~\ref{sec:eval:protocol}); the band's job is to
expose, boundary by boundary, where the model exceeds human deviation and where it does
not.
\textbf{Human disagreement predicts machine error:} the model misses existence-contested
events at roughly twice the rate of unanimous ones (gap \blMissExGap\ points,
video-clustered bootstrap 95\% CI \blMissExGapLow--\blMissExGapHigh), and chuckles at
nearly twice the rate of full laughs (gap \blMissIntGap\ points, CI
\blMissIntGapLow--\blMissIntGapHigh) - the correlated, structured failure mode
Sec.~\ref{sec:findings:chuckle} predicts, landing on the events where the single-GT
convention is least defensible. The two contrasts largely overlap (most contested events
\emph{are} chuckles, and for an audio-only model both align with the faintest acoustics),
so we count them as one confirmation, not two; but it is a direct one - the events a
real system gets wrong are disproportionately the events humans themselves contest.

\noindent\textbf{Scope.} The tier systems of Sec.~\ref{sec:eval:validation} are synthetic
(the standard device for validating a metric under a controlled quality ordering), and
Sec.~\ref{sec:eval:real} confirms the headline effects on a trained system, including
the prediction that real localizers fail on the events where human disagreement
concentrates. The recommendation stands regardless: report temporal-localization
performance against \emph{all} annotators, ideally as a distribution with conformally
calibrated tolerance, and never against a single one.
Although we instantiate it on laughter, the protocol acts only on temporal boundaries and
their inter-annotator spread, so it transfers to any affective behavior annotated as
multi-rater event segments. The nearest fit is temporal-phase segmentation of facial
expressions and action units: onset--apex--offset annotation is event-bounded in the same
shape this corpus releases, and porting needs only multi-rater phase boundaries, with the
tolerances re-calibrated per task (the offset asymmetry we measure is a property of
laughter, not of the method). Continuous dimensional traces (valence--arousal in
RECOLA~\cite{ringeval2013recola}, SEWA~\cite{kossaifi2019sewa}, and the ABAW
series~\cite{kollias2022abaw}) have no event boundaries for a band to calibrate on;
there the perspectivist argument still applies, but through amplitude- and lag-aware
mechanisms rather than a temporal tolerance. 

\section{Limitations}\label{sec:limitations}

Our analysis is based on SMILE-Temporal, whose laughter is predominantly
acoustically cued, preventing a meaningful study of audio--visual disagreement.
Furthermore, the conformal tolerance bands are calibrated to a single annotation pool and
protocol, so their widths should not be interpreted as universal. Finally, evaluation on
real systems is limited to one public audio-only laughter localizer. Extending the analysis
to additional datasets, annotation pools, and multimodal localizers remains future work.

\section{Conclusion}\label{sec:conclusion}
Single-ground-truth evaluation misrepresents temporal laughter localization. Laughter
boundaries are not point events; annotators disagree about them in structured, measurable
ways; and that disagreement should be modeled rather than averaged away. The dataset is the
enabler; the findings and the evaluation protocol are the contribution. For practitioners the prescription fits in one sentence: evaluate temporal
affect against \emph{all} annotators - ideally the distribution, with tolerance
conformally calibrated to human spread - never against a single reference; expect about
$\bandRatioNinety\times$ the tolerance at offsets as at onsets; and treat low-intensity events as intrinsically
ambiguous rather than mislabeled. We release SMILE-Disagreement as per-annotator
distributions with all analysis code - including the evaluation harness a public
laughter localizer already runs under (Sec.~\ref{sec:eval:real}), where its misses
concentrate on the events humans contest - and see two natural next steps: broadening
the real-system panel across modalities, and collecting visually-driven laughter to extend
the analysis beyond the acoustic channel.

\bibliographystyle{splncs04}
\bibliography{references}

\clearpage
\begin{center}
  {\Large\bfseries Supplementary Material}\\[2pt]
\end{center}
\bigskip

\section*{Qualitative gallery}
\begin{figure}[h]
\centering
\includegraphics[width=\textwidth]{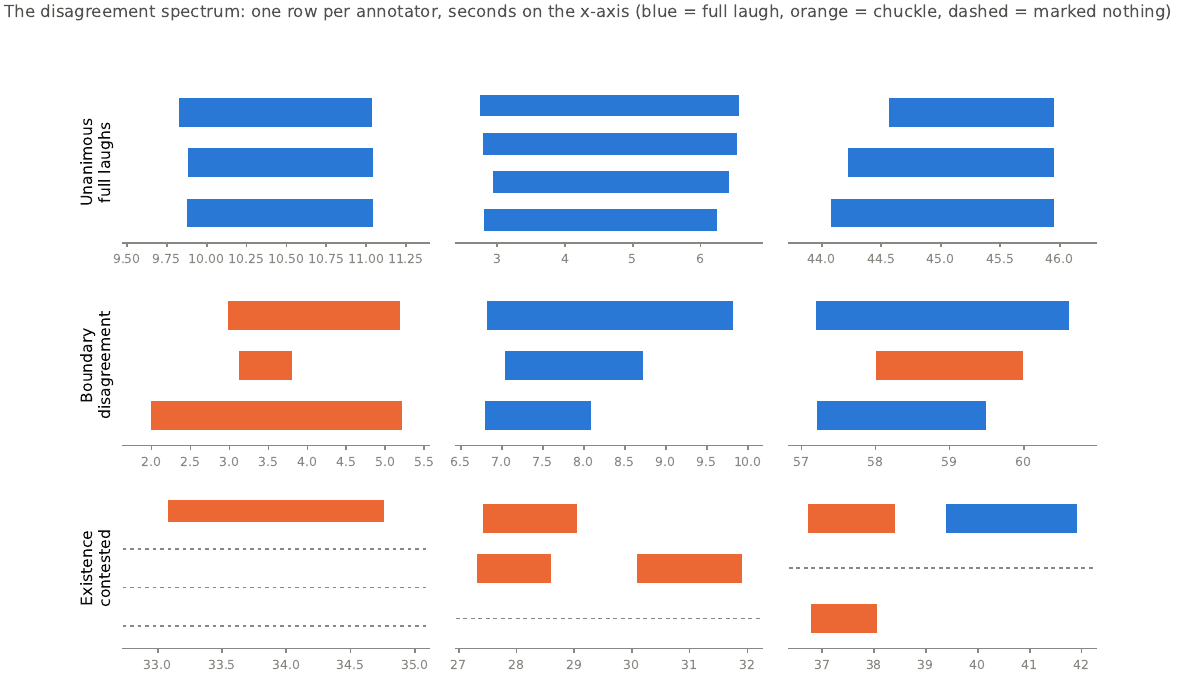}
\caption{The disagreement spectrum on nine real clips (deterministically selected; one row
per annotator; blue = full laugh, orange = chuckle, dashed line = annotator marked no
laughter). \textbf{Top:} unanimous full laughs. \textbf{Middle:} agreed events, disagreed
boundaries. \textbf{Bottom:} existence-contested clips - note they are dominated by
chuckles, the paper's headline finding.}
\label{fig:gallery}
\end{figure}

Figure~\ref{fig:gallery} shows nine real clips spanning the disagreement spectrum the paper
analyzes, in the same visual language as Fig.~1 of the main text: clear full laughs draw
near-identical intervals from every annotator; boundary-disagreement cases show the same
event with widely varying extents (typically at the offset); existence-contested cases -
overwhelmingly chuckles - split annotators on whether a laughter event is present at all.

\clearpage

\section*{Per-annotator consistency profile}
Table~\ref{tab:annotators} profiles each annotator (anonymized A1--A6, matching the
released data), supplementing Sec.~3.2 of the main paper: no annotator is an outlier, so
the disagreement structure the paper analyzes is a property of the videos, not of one
annotator's labeling style.
\begin{table}[t]
\centering
\small
\setlength{\tabcolsep}{3.5pt}
\resizebox{\columnwidth}{!}{%
\begin{tabular}{lcccccccc}
\toprule
Annotator & Videos & Ev./vid.\ & Exist.\% & Chk.\% & Aud.\% & Onset bias (s) & Offset bias (s) & F1 vs.\ peers \\
\midrule
A1 & 578 & 2.37 & 90 & 21 & 96 & -0.07 & +0.04 & 0.68 \\
A2 & 320 & 2.52 & 84 & 30 & 92 & -0.00 & -0.04 & 0.71 \\
A3 & 348 & 2.56 & 89 & 35 & 92 & +0.06 & +0.02 & 0.71 \\
A4 & 212 & 2.08 & 82 & 32 & 92 & +0.09 & -0.03 & 0.60 \\
A5 & 63 & 1.92 & 75 & 26 & 95 & +0.10 & +0.23 & 0.78 \\
A6 & 421 & 2.19 & 88 & 30 & 96 & -0.02 & -0.05 & 0.70 \\
\bottomrule
\end{tabular}%
}
\caption{Per-annotator profile on videos with at least one annotated laugh (anonymized; existence rates are thus conditional on laughter being present). Signed biases are vs.\ the median of the other annotators on events marked in common; F1 is frame-level (100\,ms) against the majority of the others; Chk.\%/Aud.\% are the chuckle and audience shares of the annotator's events (content composition is comparable across annotators). No annotator is an outlier: the findings are a property of the videos, not of one annotator's style.}
\label{tab:annotators}
\end{table}

\section*{Annotator-budget curve}
\begin{figure}[h]
\centering
\includegraphics[width=0.6\textwidth]{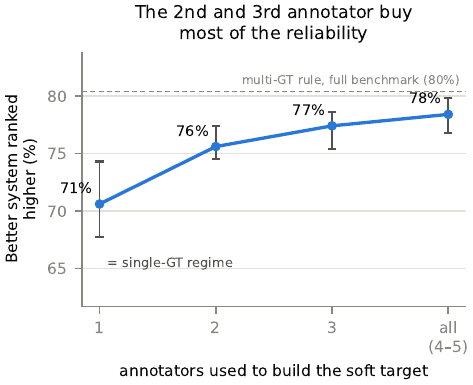}
\caption{Ranking reliability of the calibrated metric vs.\ the number of annotators used to
build the soft target (${\geq}4$-annotator videos; whiskers: min--max over seeds). The
second and third annotator buy most of the reliability - practical guidance for
annotation-budget planning.}
\label{fig:budget}
\end{figure}

Figure~\ref{fig:budget} plots the ranking-reliability-vs-annotator-count result quoted in
Sec.~5.3 of the main paper.
\clearpage

\section*{Summary of findings}
Table~\ref{tab:findings} collects the main-paper findings (Sec.~4) and their robustness
checks in one place; every value also appears in the Sec.~4 prose.
\begin{table}[h]
\centering
\small
\setlength{\tabcolsep}{3pt}
\resizebox{\textwidth}{!}{%
\begin{tabular}{lcc}
\toprule
Measure & Value & Test \\
\midrule
Frame agreement $\alpha$ (100\,ms)      & \frameAlpha\ [\bootAlphaLow, \bootAlphaHigh] & - \\
\quad across 40--500\,ms bins           & \alphaBinLow--\alphaBinHigh & stable \\
\quad subsampled to 3 annotators        & \frameAlphaSub & stable \\
Existence unanimous / contested         & \existUnanimous\% / \existContested\% & - \\
Onset s.d.\ (matched events)            & \onsetSD\,s & \multirow{2}{*}{$p{=}\asymWilcoxonP$} \\
Offset s.d.\ (matched events)           & \offsetSD\,s & \\
Offset/onset spread ratio               & \asymRatioLow--\asymRatioHigh$\times$ & range over 3 matchers \\
\quad within audience / indiv.-speaker events & \asymRatioAud$\times$ / \asymRatioSpk$\times$ & (confound ctrl.) \\
\quad acoustic decay/attack (RMS envelope) & \decayOverAttack$\times$ & CI [\decayOverAttackLow, \decayOverAttackHigh] \\
Disagreement rate: chuckle vs.\ full    & \chuckleDisagree\% vs.\ \fullDisagree\% & gap [\bootGapLow, \bootGapHigh] pts \\
\quad holds within acoustic-only events & yes & (confound ctrl.) \\
\quad holds within duration quartiles   & gap \chuckleGapStratMin--\chuckleGapStratMax\,pts & (confound ctrl.) \\
\quad at the 3-annotator floor          & \chuckleDisagreeSub\% vs.\ \fullDisagreeSub\% & stable \\
\quad across event matchers             & \chuckleMatchLow--\chuckleMatchHigh\% vs.\ \fullMatchLow--\fullMatchHigh\% & stable \\
\quad intensity-unanimous events only   & \chuckleDisagreeUnanInt\% vs.\ \fullDisagreeUnanInt\% & stable \\
Boundary spread: chuckle vs.\ full (offset) & \chOffsetSD\ vs.\ \fullOffsetSD\,s & \bndQuartRatioLow--\bndQuartRatioHigh$\times$ per duration quartile \\
\bottomrule
\end{tabular}%
}
\caption{Summary of findings on the \nBenchmark-video benchmark. Brackets are video-level
clustered bootstrap 95\% CIs ($B{=}1000$); the onset/offset test is a one-sided Wilcoxon
signed-rank test over matched events.}
\label{tab:findings}
\end{table}

\section*{Protocol illustration}
\begin{figure}[h]
\centering
\includegraphics[width=0.62\textwidth]{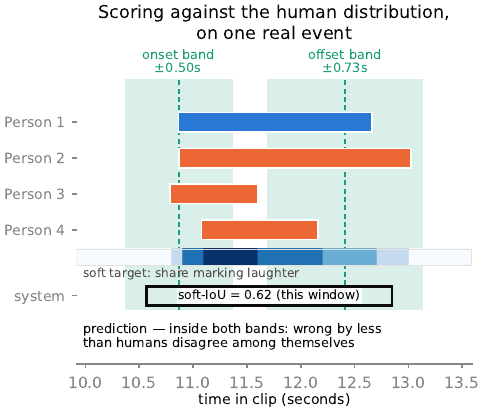}
\caption{The protocol on a real event from the corpus. Four annotators mark the same laugh
(top rows; blue = full laugh, orange = chuckle); their per-frame fraction is the \emph{soft
target} (shaded strip); the conformally calibrated tolerance bands (teal, drawn to scale)
are exactly as wide as human disagreement - $\pm\bandOnsetNinety$\,s at the onset but
$\pm\bandOffsetNinety$\,s at the offset. The example prediction (black outline; its
soft-IoU restricted to the displayed window, matching the on-figure label) deviates from
the median annotator yet stays within both bands:
it is wrong only by less than humans disagree among themselves.}
\label{fig:protocol}
\end{figure}

Figure~\ref{fig:protocol} illustrates the disagreement-calibrated protocol (Sec.~5.2 of the
main paper) on a single real event.

\section*{Conformal band: specification details}
Three details supplement the band's definition in Sec.~5.2 of the main paper.

\emph{Calibration/test $k$-mismatch.} Calibration holds out one annotator (median over
$k{-}1$) while a prediction is tested against the median of all $k$ - the more stable of
the two, so the applied band errs conservative.

\emph{Per-co-marker-count strata.} The pooled quantile is not an artifact of the thinnest
events. Splitting the leave-one-annotator-out calibration pool by co-marker count: with
$k{=}2$ co-markers ($n{=}\nBandKTwo$ residuals per boundary type) the 90\%-coverage
half-widths are $\pm\bandOnKTwo$\,s (onset) and $\pm\bandOffKTwo$\,s (offset); with
$k{\geq}3$ ($n{=}\nBandKThree$) they are $\pm\bandOnKThree$\,s and $\pm\bandOffKThree$\,s
- close to the pooled $\pm\bandOnsetNinety$/$\pm\bandOffsetNinety$\,s, so the pooled
quantile is not driven by the thin-calibration events.

\emph{Adaptive variant.} The adaptive band normalizes each nonconformity score $r$ by the
local spread $\sigma_e$ of the co-markers' boundaries (floored at $0.05$\,s), giving a
per-event half-width $\tilde{q}_b\,\sigma_e$ with the same 90\% coverage and a median
width of $\pm\adBandOnsetNinety$\,s at onsets and $\pm\adBandOffsetNinety$\,s at offsets
(vs.\ the global $\pm\bandOnsetNinety$/$\pm\bandOffsetNinety$\,s) - tight where
annotators agree, wide only where they do not.
\clearpage

\end{document}